\documentclass[10pt,twocolumn,letterpaper]{article}

\usepackage[pagenumbers]{cvpr} 

\definecolor{cvprblue}{rgb}{0.21,0.49,0.74}
\usepackage[pagebackref,breaklinks,colorlinks,allcolors=cvprblue]{hyperref}

\def\paperID{***} 
\def\confName{CVPR}
\def\confYear{2025}

\usepackage{hyperref}
\usepackage{graphicx}
\usepackage{multirow}

\usepackage{algorithm}
\usepackage{alphalph}
\usepackage{algorithmic}
\usepackage{bbding}
\usepackage{makecell}
\usepackage[accsupp]{axessibility}

\usepackage{orcidlink}

\usepackage{enumitem}

\usepackage[utf8]{inputenc} 
\usepackage[T1]{fontenc}    
\usepackage{url}            
\usepackage{booktabs}       
\usepackage{amsfonts}       
\usepackage{nicefrac}       
\usepackage{microtype}      
\usepackage{amssymb}
\usepackage{pifont}

\newcommand{\name}{InstructSeg}
\newcommand{\dname}{Unifying Instructed Visual Segmentation \\ with Multi-modal Large Language Models}

\title{\name: \dname}

\author{Cong Wei$^{1,2}$, Yujie Zhong$^{2}$$^{\dagger}$, Haoxian Tan$^{2}$, Yingsen Zeng$^{2}$, Yong Liu$^{1}$, Zheng Zhao$^2$, and Yujiu Yang$^{1}$$^{\dagger}$  \\
$^1$Tsinghua Shenzhen International Graduate School, Tsinghua University 
   $^2$Meituan Inc.\\
{\tt\small weic22@mails.tsinghua.edu.cn,} 
{\tt\small jaszhong@hotmail.com,} 
{\tt\small yang.yujiu@sz.tsinghua.edu.cn}
}

\begin{document}

\twocolumn[{%
\renewcommand\twocolumn[1][]{#1}%
\maketitle
\vspace{-9mm}
\centering
\includegraphics[width=\textwidth]{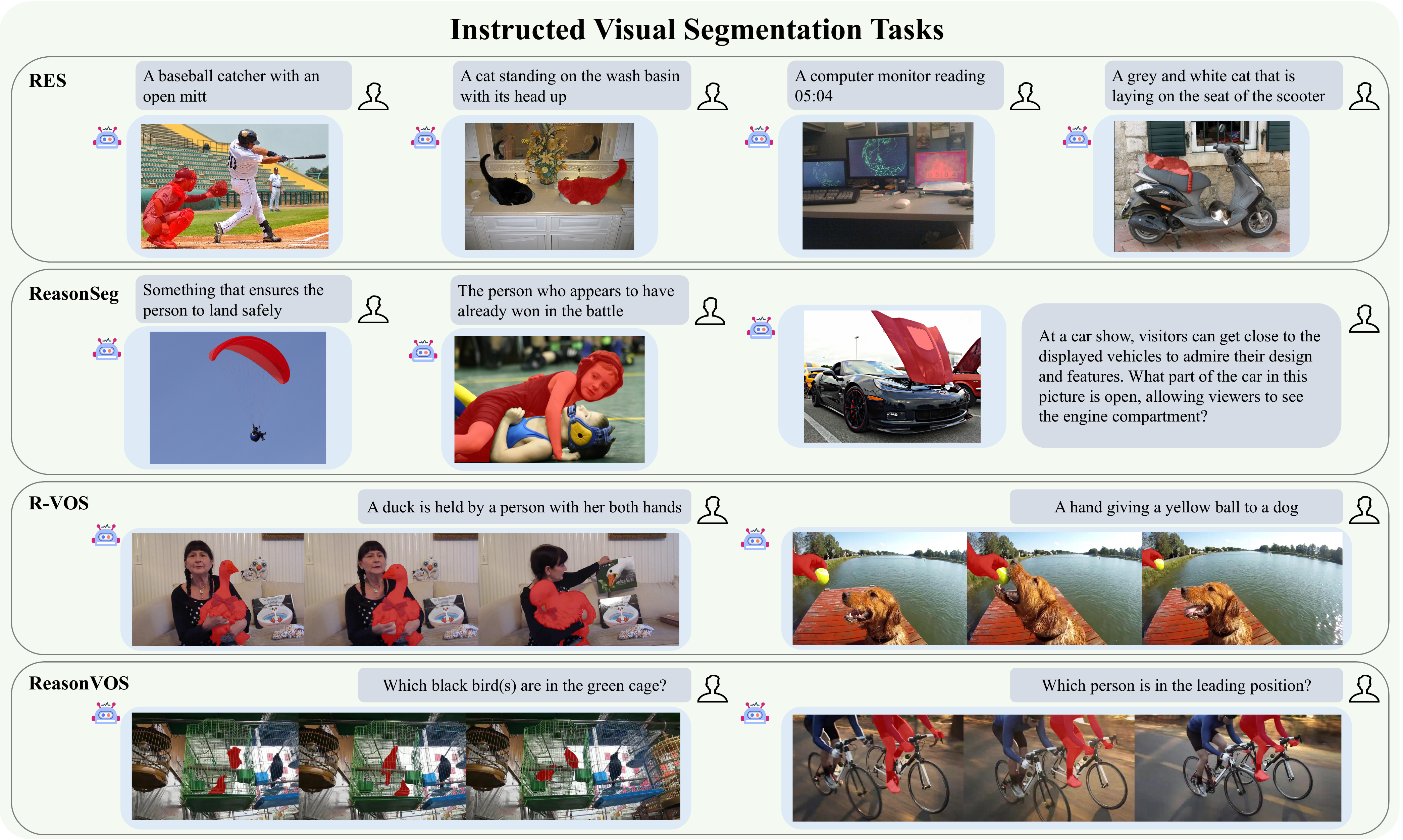}
\vspace{-4mm}
\captionof{figure}{We define Instructed Visual Segmentation (IVS) as the union of four text-guided segmentation tasks across image and video domains: referring expression segmentation (RES), reasoning segmentation (ReasonSeg), referring video object segmentation (R-VOS) and reasoning video object segmentation (ReasonVOS). \name~can handle all the IVS tasks in one model with excellent performance.}
\label{fig:intro}
\vspace{0.6cm}
}]

\renewcommand{\thefootnote}{\fnsymbol{footnote}}
\footnotetext{$^{\dagger}$Corresponding authors.}

\begin{abstract}
\indent 
Boosted by Multi-modal Large Language Models (MLLMs), text-guided universal segmentation models for the image and video domains have made rapid progress recently. However, these methods are often developed separately for specific domains, overlooking the similarities in task settings and solutions across these two areas.
In this paper, we define the union of referring segmentation and reasoning segmentation at both the image and video levels as Instructed Visual Segmentation (IVS). Correspondingly, we propose \name, an end-to-end segmentation pipeline equipped with MLLMs for IVS.
Specifically, we employ an object-aware video perceiver to extract temporal and object information from reference frames, facilitating comprehensive video understanding. Additionally, we introduce vision-guided multi-granularity text fusion to better integrate global and detailed text information with fine-grained visual guidance. By leveraging multi-task and end-to-end training, \name~demonstrates superior performance across diverse image and video segmentation tasks, surpassing both segmentation specialists and MLLM-based methods with a single model. Our code is available \href{https://github.com/congvvc/InstructSeg}{here}.
\vspace{-4mm}

\end{abstract}

\begin{table*}[t]
  \centering
  \caption{ The comparison of capabilities of different methods. Our \name~can tackles more comprehensive Instructed Visual Segmentation tasks with a simplified end-to-end framework and achieves better performance.
  }
  \scalebox{0.88}{
    \begin{tabular}{lccccc}
    \toprule[1.1pt] 
     \multirow{2}{*}{ Method } & \multirow{2}{*}{ End-to-End } & \multicolumn{4}{c}{ Instructed Visual Segmentation } \\
     & & RES & ReasonSeg & R-VOS & ReasonVOS  \\
    \midrule[0.5pt]
    \emph{Segmentation Specialists} \\
    \midrule[0.5pt] 
    
    VLT~\cite{ding2021vision}, LAVT~\cite{yang2022lavt}, ReLA~\cite{liu2023gres}, PolyFormer~\cite{liu2023polyformer} & \Checkmark & \Checkmark &  &  &   \\
    ReferFormer~\cite{wu2022language}, OnlineRefer~\cite{wu2023onlinerefer}, SOC~\cite{luo2024soc}& \Checkmark &  &  & \Checkmark &   \\
    
    \midrule[0.5pt] 
    \emph{MLLM-based Generalists
    } \\
    \midrule[0.5pt] 
    LISA~\cite{Lai2023LISARS}, PixelLM~\cite{Ren2023PixelLMPR}, LaSagnA~\cite{wei2024lasagna} & \Checkmark & \Checkmark & \Checkmark &  &  \\
    GSVA~\cite{xia2023gsva}, GLaMM~\cite{glamm}, OMG-LLaVA ~\cite{zhang2024omg}, PSALM ~\cite{zhang2024psalm} & \Checkmark & \Checkmark &  &  &  \\
     VISA ~\cite{yan2024visa} &  & \Checkmark & \Checkmark &  \Checkmark &  \Checkmark \\
    
    \name~(Ours) & \Checkmark & \Checkmark & \Checkmark &  \Checkmark &  \Checkmark \\

    \bottomrule[1.1pt]
    \end{tabular}
}

\label{tab:task-compare}
\end{table*}

\section{Introduction}\label{sec:intro}

In recent years, language-instructed segmentation tasks, such as referring expression segmentation (RES) \cite{ding2021vision, wang2022cris, yang2022lavt, liu2023polyformer} and referring video object segmentation (R-VOS) \cite{ding2023mevis, wu2022language, wu2023onlinerefer}, have attracted significant attention in the field of computer vision. These tasks require models to segment objects in images or videos based on natural language descriptions, offering greater potential for practical applications but also presenting a more substantial challenge compared to traditional class-based segmentation tasks like semantic segmentation, which depend on predefined categories. Multi-modal Large Language Models (MLLMs) are distinguished by their ability to handle versatile inputs and their impressive performance in visual-language reasoning, making them a preferred choice for these language-instructed segmentation tasks in both image and video domains. Despite the similarity in task settings and solutions, current research often overlooks the collaborative effect of these tasks in the context of multi-task optimization, treating them separately.

Recent works in the image domain, such as LISA \cite{Lai2023LISARS}, PixelLM \cite{Ren2023PixelLMPR}, and PSALM \cite{zhang2024psalm}, have demonstrated promising results in universal image segmentation by leveraging MLLMs' embedded reasoning capabilities and inherent world knowledge. However, models designed for static images typically \textbf{fail to effectively capture the temporal coherence within videos}, either by naively aggregating video frames or by processing video frames individually, thereby incurring excessive computational costs.

Meanwhile, recent work in the video domain, VISA~\cite{yan2024visa}, has sought to address these challenges by leveraging the combination of multiple specialists for text-guided image and video segmentation. 
Nonetheless, VISA requires the integration of each specialist, necessitating the use of additional pre-trained Video-LLMs and a video object tracker. The complex model and cumbersome pipeline impede the potential for widespread application and \textbf{end-to-end collaborative optimization in both image and video domains.}

In this work, we unify referring segmentation and reasoning segmentation within both the image and video domains under the framework of \textbf{Instructed Visual Segmentation (IVS)}. IVS requires models to segment targets from visual inputs based on detailed textual instructions. As shown in \cref{tab:task-compare}, IVS encompasses traditional referring segmentation tasks, such as referring expression segmentation (RES) and referring video object segmentation (R-VOS), as well as more complex reasoning-based segmentation tasks, including reasoning segmentation (ReasonSeg) and reasoning video object segmentation (ReasonVOS).

Accordingly, we introduce an end-to-end pipeline, \name, along with two meticulous designs: Object-aware Video Perceiver and Vision-guided Multi-granularity Text Fusion. This pipeline enables the implementation of all tasks within IVS using a single MLLM and segmentation decoder. 

Specifically, Object-aware Video Perceiver is designed to extract both the temporal and object information from reference frames guided by language instructions referring to or reasoning about the objects of interest.
Vision-guided Multi-granularity Text Fusion is proposed to comprehend long text instructions and complex scenarios like “A bride and groom often walk together down the aisle during a wedding ceremony. What object in the picture is the bride most likely holding during this moment?”. Instead of averaging or summarizing the embeddings of multiple text tokens, we integrate both global and detailed information of text instructions into the multi-granularity text embeddings, facilitating better comprehension of the intentions behind the text instructions.

Extensive experiments on various  Instructed Visual Segmentation benchmarks demonstrate the powerful reasoning and segmentation abilities of our \name.
Our contributions are as follows: 
\begin{itemize}[leftmargin=0.5cm]
   
    \item We integrate referring segmentation and reasoning segmentation across both image and video domains within the framework of Instructed Visual Segmentation (IVS). Introducing \name, we offer an end-to-end unified pipeline for all IVS tasks, leveraging Multi-modal Large Language Models for language-instructed pixel-level reasoning and classification. This model effectively handles tasks across both image and video domains while maintaining a high level of simplicity.
    \item We propose Object-aware Video Perceiver and Vision-guided Multi-granularity Text Fusion modules to fully exploit temporal and object information, enhancing the understanding of both global and detailed text instructions.
    \item \name~achieves state-of-the-art results on diverse Instructed Visual Segmentation benchmarks across both image and video domains, demonstrating the effectiveness of our simplified pipeline. Additionally, \name~delivers competitive performance on multiple Multi-modal benchmarks.
    
\end{itemize}

\section{Related Work}
\textbf{Referring segmentation in image and video.} The referring expression segmentation (RES) task aims at labeling the pixels of an image that represent an object instance referred to by a linguistic expression. CRIS~\cite{wang2022cris}, CGFormer~\cite{tang2023contrastive}, CoupAlign~\cite{zhang2022coupalign} firstly use visual language decoder to propagate text information to pixel-level visual features, and then bring relevant features closer based of contrastive Learning. MCN~\cite{luo2020multi}, RefTR~\cite{li2021referring}, and X-Decoder~\cite{zou2023generalized} enhance the fusion of textual and visual features and utilize multi-task collaborative learning to strengthen the segmentation performance. 
The referring video object segmentation (R-VOS) extends the RES task to the domain of videos, where the goal is to track and segment the object corresponding to a given natural language description in a video. 
The method ~\cite{bellver2023closer} extracts frames from the video and processes them separately. 
However, this method disregards the temporal information of the video, making it susceptible to challenges such as object motion and lighting variations. 
On the other hand, the methods \cite{kamath2021mdetr, luo2020multi, botach2022end, wu2022language} detect and propagate the target masks. Both RES and R-VOS are based on explicit instructions but lack the requirements to reason about complex tasks. Recently proposed ReasonSeg~\cite{Lai2023LISARS} and ReasonVOS~\cite{yan2024visa} tasks aim to address this limitation, which extends short phrases to complex sentences that require reasoning and the inference of world knowledge in conjunction with video content.

\noindent \textbf{Multi-modal Large Language Models.} Inspired by the remarkable comprehension and reasoning capabilities of large language models, considerable effort have been made within both the open-source community and research institutions to expand these models to encompass various modalities, resulting in the development of MLLMs. For instance, Flamingo~\cite{alayrac2022flamingo} incorporates a perceiver resampler to convert visual information and employs a gated cross-attention layer to establish deeper feature fusion between the frozen visual encoder and the LLM. LLAVA~\cite{liu2024visual}, BLIP2~\cite{li2023blip} and miniGPT4~\cite{zhu2023minigpt} primarily use vision encoders to encode images and incorporate modal adaptors to map visual features into the text domain, serving as the input for LLMs. Additionally, Video-Chat~\cite{li2023videochat} and Video-ChatGPT~\cite{maaz2023video} extend image encoders to video encoders, facilitating the understanding of visual content in videos. 
Models such as PLLaVA~\cite{xu2024pllava} and MiniGPT4-video~\cite{ataallah2024minigpt4} process each frame independently using an image encoder and reduce the number of video visual tokens through temporal dimension pooling.
Blip3-video~\cite{ryoo2024xgen} introduces a temporal encoder to map a sequence of tokens from multiple frames into a compact set of visual tokens. 
Furthermore, LLaMA-VID~\cite{li2025llama} even represents a frame with only two tokens, making it possible to handle inference on long videos lasting over an hour. However, these methods are primarily designed for tasks that require text output, like VQA, and are not directly applicable to pixel-level understanding tasks like image segmentation and video segmentation.

\noindent \textbf{MLLM-based segmentation model.} Several existing works aim to utilize MLLMs to incorporate complex reasoning and world knowledge, as well as enable MLLMs to generate segmentation masks in images. These methods including LISA~\cite{Lai2023LISARS}, PerceptionGPT~\cite{pi2023perceptiongpt}, GSVA~\cite{xia2023gsva}, PixelLM~\cite{Ren2023PixelLMPR}, LaSagnA~\cite{wei2024lasagna}, PSALM~\cite{zhang2024psalm}, HyperSeg~\cite{wei2024hyperseg} \etc, employ a common approach of introducing a segment token for each sentence related to a different object. Specifically, LISA~\cite{Lai2023LISARS}, PerceptionGPT~\cite{pi2023perceptiongpt} and GSVA~\cite{xia2023gsva} employ the segment token from the MLLM as a prompt embedding and further employ the SAM~\cite{kirillov2023segment} model to generate segmentation predictions. PixelLM~\cite{Ren2023PixelLMPR} generates multi-scale segment tokens from the MLLM and a segmentation codebook, which are subsequently processed by a lightweight decoder to produce weighted masks. Conversely, PSALM~\cite{zhang2024psalm} follows Mask2Former~\cite{cheng2022masked}, generating mask proposals first and then classifying them instead of directly generating the final prediction. However, these models are limited to static image predictions and do not effectively leverage temporal information for video segmentation tasks. To address video tasks, VISA~\cite{yan2024visa} employs a frame sampler to selectively choose the most relevant frames to the textual instructions. Then the visual token and text sentences are jointly processed using MLLM to derive reasoning over the video content and generate precision textual outputs. Finally, the video object segment output is obtained by the SAM mask decoder and X-Mem~\cite{cheng2022xmem} tracker. Although VISA~\cite{yan2024visa} supports segment tasks for both images and videos, it relies on the pre-trained model quality for the frame sampler and tracker. These modules are frozen during training, hindering the synergistic optimization of perception and segmentation. 
To this end, we propose an end-to-end framework that unifies reasoning and segmentation into a single module, allowing for simultaneous optimization and improved performance.

\begin{figure*}[t]
    \centering
    \includegraphics[width=\textwidth]{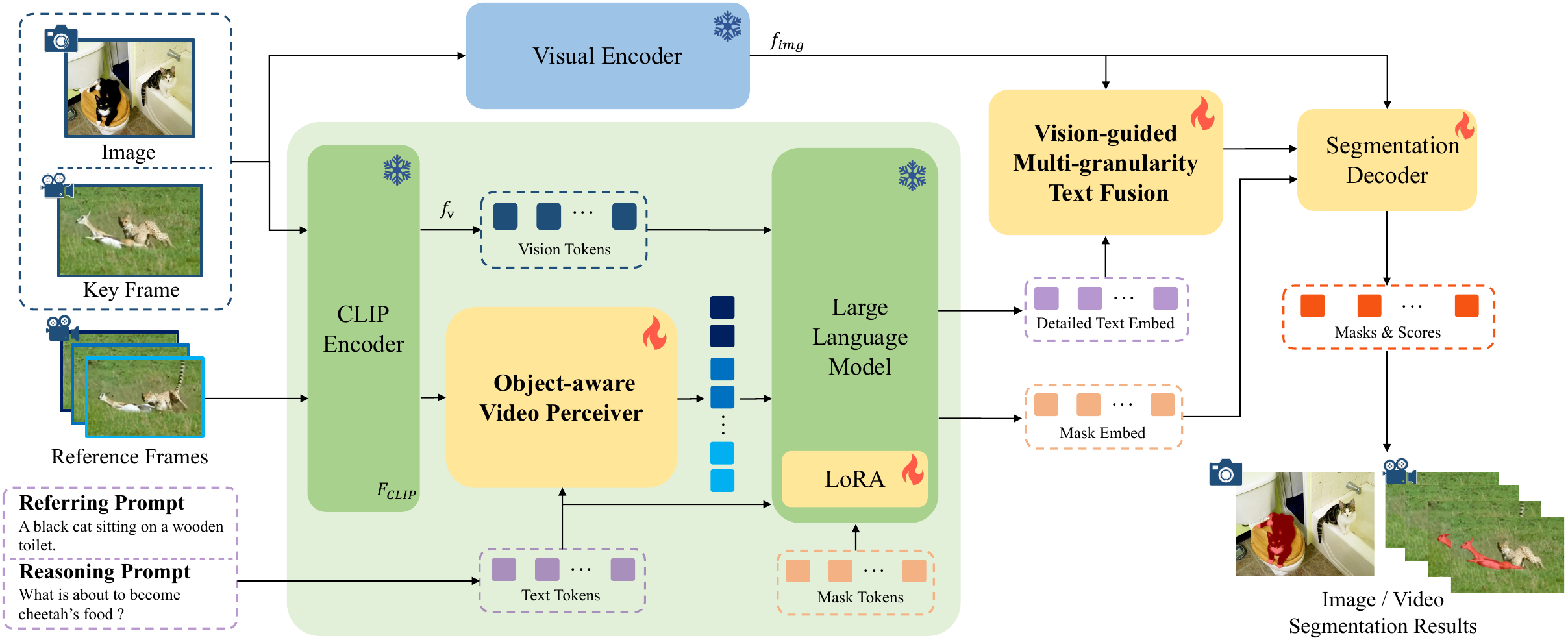}
    \caption{Framework of \name. \name~tackles Instructed Visual Segmentation tasks in an end-to-end pipeline.
    For challenging video analysis tasks, we employ the object-aware video perceiver to effectively extract both temporal and object-specific information from the reference frames. 
    Besides, \name~is capable of executing comprehensive and accurate vision-language perception and understanding 
    through vision-guided multi-granularity text fusion applied to detailed text embeddings.
    Finally, the mask embeddings and multi-granularity text embeddings are decoded into segmentation masks and scores.}

    \label{fig:model}
\end{figure*}

\section{Method} \label{sec:method}

\subsection{Overview of \name}
\label{subsec:overview}

The architecture of our proposed model, referred to as \name, is depicted in \cref{fig:model}. This model primarily comprises several key components: a Multi-modal Large Language Model (integrating a CLIP encoder and a Large Language Model), a visual encoder, an object-aware video perceiver (OVP), a vision-guided multi-granularity text fusion module (VMTF), and a segmentation decoder responsible for generating masks and scores.
The model processes several inputs: an image or key frame denoted as $\mathcal{V} \in \mathbb{R}^{H\times W\times 3}$, a reference video sequence  $\mathcal{V}_r \in \mathbb{R}^{T_r\times H\times W\times 3}$,  comprising $T_r$ frames for video tasks, and language instructions $\mathcal{E}$. The CLIP encoder is tasked with encoding all visual inputs into global image features. Concurrently, the object-aware video perceiver handles the reference frames and text tokens, utilizing learnable queries to condense temporal and object information into fixed-length tokens. The Large Language Model (LLM) processes four distinct types of inputs: visual tokens derived from the image or key frame, text tokens, compressed tokens from the reference frames, and mask tokens. The output embeddings of the LLM, which correspond to detailed text instructions, are blended using the VMTF module under visual guidance. Subsequently, the segmentation decoder produces segmentation masks and scores by decoding the mask embeddings and multi-granularity text embeddings from the VMTF output, alongside fine-grained visual features $f_{img}$. 
It is important to note that all visual encoders are frozen, while the remaining components are trained. Specifically, the LLM is fine-tuned using Low-Rank Adaptation (LoRA) to enhance tuning efficiency.

\subsection{Object-aware Video Perceiver}
\label{subsec:ovp}
Video segmentation poses distinct challenges, primarily due to the need for reasoning across multiple frames while ensuring temporal coherence. Recent models, such as VISA~\cite{yan2024visa}, employ pre-trained video specialists to tackle these challenges. However, this approach often results in a heavy dependence on the capabilities of individual components, which can lead to error accumulation throughout the process.
In response to these challenges, we introduce the Object-aware Video Perceiver (OVP). The OVP module is designed to extract both temporal and object-specific information from reference frames, guided by language instructions that are expected to identify the objects of interest. This approach aims to enhance the accuracy and coherence of video segmentation by effectively integrating temporal and contextual information.

\begin{figure}[ht]
    \centering
    \includegraphics[width=0.46\textwidth]{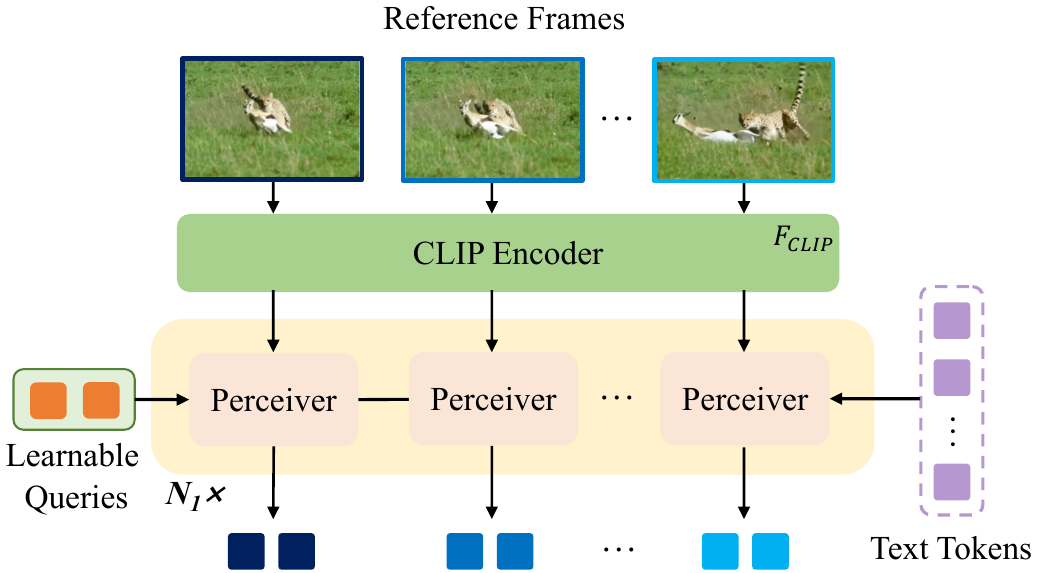}
    \caption{Illustration of Object-aware Video Perceiver (OVP). OVP learns temporal and object information with $N_1$ perceiver layers through the interactions of text and reference frames along with the learnable queries.}
    \label{fig:ovp}
    \vspace{-3mm}
\end{figure}

As shown in \cref{fig:ovp}, given the $T_r$ reference frames $\mathcal{V}_r = \{I_r^{t}\}_{t=1}^{T_r}$ alongside text tokens $\mathcal{E}$, we initially utilize the CLIP encoder, denoted as $F_{CLIP}$, to extract frame-level visual features $f_r^t$. Subsequently, we employ $N_1$  perceiver layers to integrate vision-language information into learnable queries $Q$ for each reference frame.
Formally, 
\begin{equation}
\begin{aligned}
    f_{r}^t&=F_{CLIP}(I_r^{t}), \\
    Q^t&=\textrm{CrossAttn}(Q, Concat[F_{p}(f_{r}^{t}), \mathcal{E}]).
\end{aligned}
\label{Eq: ovp}
\end{equation}
where $\textrm{CrossAttn}$ denotes the Multi-Head Cross-Attention layer in perceiver, $F_p$ is the projection function to align vision-language space, 
$Q^t$ denotes the output reference tokens for the $t$-th reference frame.

Finally, we concat all the reference tokens $Q^t$ along the time dimension to get the general reference tokens $Q_{r}$.
We feed the general reference tokens $Q_{r}$ into LLM along with text and visual tokens to tackle complex video reasoning tasks and temporal coherence problems. The implementation details will be elaborated upon in the following sections. Besides, for image segmentation tasks, the image itself serves as the reference for our OVP module, facilitating the bridge of the gap between image and video tasks (demonstrated in Tab.~\ref{tab:ab-component}).

\noindent \textbf{Co-Decoding Process of LLM.}
We utilize LLM for the collaborative decoding process of multiple functional tokens, including visual tokens and text tokens for common Multi-modal comprehension, general reference tokens for video reasoning and understanding, and mask tokens for generating segmentation masks.
To be specific, given the image or key frame $\mathcal{V}$, general reference tokens $Q_{r}$ from OVP module for video tasks, text tokens $\mathcal{E}$, and initialized mask tokens $M$. Large Language Model $F_{LLM}$ integrates them together and outputs the corresponding embeddings for the following multi-granularity text fusion and mask decoding process. 
Formally,
\begin{equation}
     f_{v}=F_{CLIP}(\mathcal{V}), E=F_{LLM}(F_{p}(f_{v}), Q_{r}, \mathcal{E}, M).
\label{Eq: llm}
\end{equation}
where $F_{p}$ is the projection function and $E$ denotes the output embeddings of LLM.
Subsequently, we extract the detailed text embeddings $E_d$ and mask embeddings $E_m$ from $E$ according to their respective indexes, which are then fed into the following VMTF module and segmentation decoder for multi-granularity text fusion and mask generation.

\subsection{Vision-guided Multi-granularity Text Fusion }
\label{subsec:vmtf}

In the context of Instructed Visual Segmentation tasks, particularly reasoning-based segmentation, it is critical to infer the truly desired or interesting objects from detailed textual descriptions. Previous methods~\cite{Lai2023LISARS,zhang2024psalm,yan2024visa} typically average or summarize the embeddings of multiple text tokens to generate a global embedding as the segmentation mask classifier. They tend to overlook detailed text information, which is crucial for understanding complex scenarios. Therefore, we propose the Vision-guided Multi-granularity Text Fusion (VMTF) module to integrate both global and detailed information of language instructions into the multi-granularity text embeddings as the mask classifier, facilitating better comprehension of the intentions behind the text instructions.

\begin{figure}[t]
    \centering
    \includegraphics[width=0.48\textwidth]{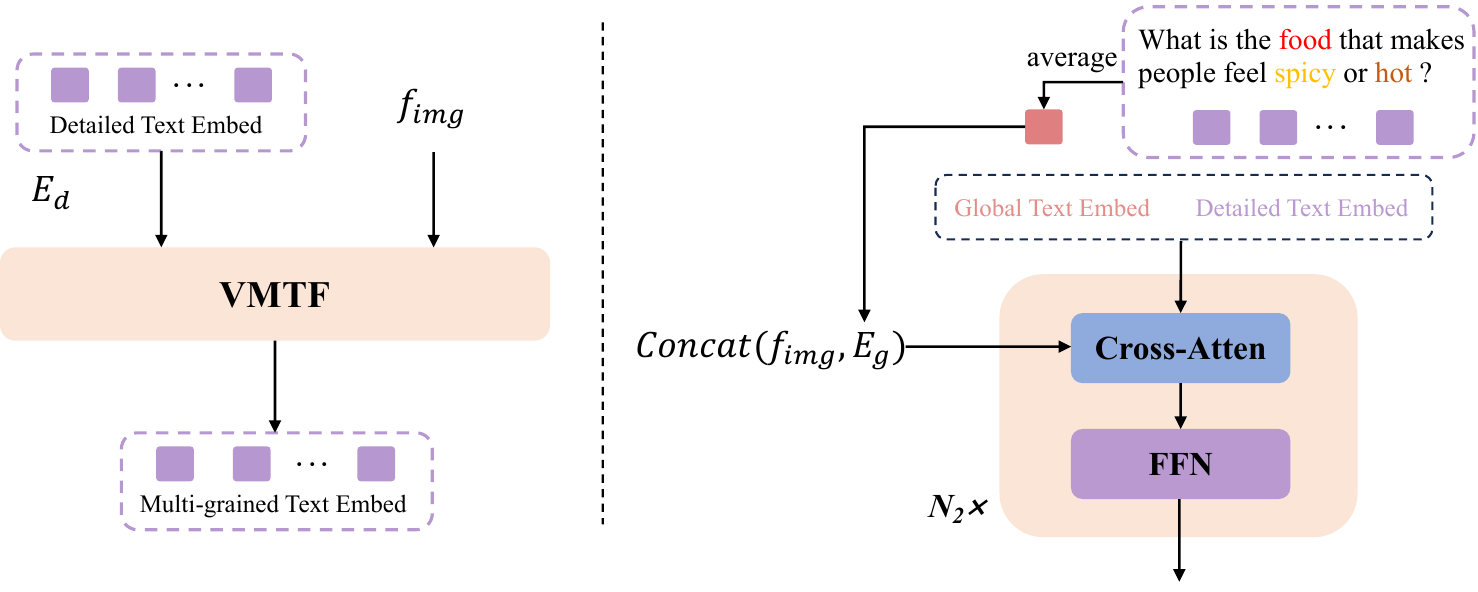}
    \caption{The structure of the Vision-guided Multi-granularity Text Fusion (VMTF) module.}
    \label{fig:vmtf}
    \vspace{-3mm}
\end{figure}

As shown in \cref{fig:vmtf}, given the detailed text embeddings $E_d$, we initially employ an adaptive average pooling strategy to obtain global text embeddings $E_g$, which consolidate global textual information. Subsequently, both the global and detailed text embeddings are input into $N_2$ cross-attention and FFN layers along with the concatenation of fine-grained image features $f_{img}$ and global text embeddings $E_g$.
The Vision-guided Multi-granularity Text Fusion (VMTF) module is then utilized to integrate global and detailed language instruction information into multi-granularity text embeddings $E_{v}$. This integration enhances the accuracy and robustness of mask classification in both image and video domains.

\subsection{Mask Decoding and Training objectives}
\label{subsec:decoding}

After the collaborative decoding and comprehension of LLM on multiple functional tokens, the segmentation decoder generates more accurate object masks and scores under the guidance of multi-granularity text embeddings.

\begin{table*}[t]
  \centering
  \caption{ Comparison with the state-of-the-art models on the referring expression segmentation benchmarks (refCOCO/+/g) and reasoning segmentation (ReasonSeg). Our \name~exhibits superior performance, surpassing all other methods significantly including both segmentation specialists and MLLM-based methods.
  }
  \scalebox{0.9}{
    \begin{tabular}{c|l|ccc|ccc|cc|cc}
    \toprule[1.1pt] 
    \multirow{2}{*}{ Type } & \multirow{2}{*}{ Method } & \multicolumn{3}{c|}{ refCOCO } & \multicolumn{3}{c|}{ refCOCO+ } & \multicolumn{2}{c|}{ refCOCOg } & \multicolumn{2}{c}{ ReasonSeg }\\
    \cline { 3 - 12 } & & val & testA & testB & val & testA & testB & val(U) & test(U) & gIoU & cIoU\\
    \midrule[0.7pt]
    \multirow{6}{*}{\shortstack{Segmentation\\Specialists}} 
    & VLT~\cite{ding2021vision} & 67.5 & 70.5 & 65.2 & 56.3 & 61.0 & 50.1 & 55.0 & 57.7  & - & -\\
    & CRIS~\cite{wang2022cris} & 70.5 & 73.2 & 66.1 & 62.3 & 68.1 & 53.7 & 59.9 & 60.4  & - & -\\
    & LAVT~\cite{yang2022lavt} & 72.7 & 75.8 & 68.8 & 62.1 & 68.4 & 55.1 & 61.2 & 62.1  & - & -\\
    & ReLA~\cite{liu2023gres}  & 73.8 & 76.5 & 70.2 & 66.0 & 71.0 & 57.7 & 65.0 & 66.0 & - & -\\
    & PolyFormer~\cite{liu2023polyformer} & 74.8 & 76.6 & 71.1 & 67.6 & 72.9 & 59.3 &67.8 &69.1 & - & -\\
    & SEEM~\cite{zou2024segment} & - & - & - & - &- & - &67.7 &- & 25.5 & 21.2\\
    \midrule[0.5pt] 
    \multirow{10}{*}{\shortstack{MLLM-based\\Segmentation Models}} & LISA-7B~\cite{Lai2023LISARS} & 74.9 & 79.1 & 72.3 & 65.1 & 70.8 & 58.1 &  67.9 & 70.6  & 52.9 & 54.0\\
    & PixelLM-7B~\cite{Ren2023PixelLMPR}& 73.0 &76.5 & 68.2 & 66.3 & 71.7 & 58.3 &69.3 & 70.5  & - & -\\
    & GSVA-7B~\cite{xia2023gsva}& 76.4 &77.4 &72.8 &64.5 &67.7 & 58.6 & 71.1 & 72.0  & - & -\\
    & LaSagnA-7B~\cite{wei2024lasagna} & 76.8 & 78.7 & 73.8 & 66.4 & 70.6 & 60.1 & 70.6 & 71.9  & 48.8 & 47.2\\
    &VISA-7B ~\cite{yan2024visa} &72.4 &75.5 &68.1 &59.8 &64.8 &53.1 &65.5 &66.4 & 52.7 &57.8 \\
    & OMG-LLaVA ~\cite{zhang2024omg} & 78.0   & 80.3  & 74.1   & 69.1    & 73.1     & 63.0   & 72.9     & 72.9   & - & -\\
    & GroundHog~\cite{miao2023spectrum} & 78.5 & 79.9 & 75.7 & 70.5 & 75.0 & 64.9 & 74.1 & 74.6 & 56.2 & - \\
    &GLaMM~\cite{glamm} & 79.5    & 83.2    & 76.9  & 72.6    & 78.7     & 64.6    &  74.2   & 74.9   & - & -  \\
    & PSALM ~\cite{zhang2024psalm} & 83.6   & 84.7  & 81.6   & 72.9    & 75.5     & 70.1   & 73.8     & 74.4   & - & -\\ 
    
    & \textbf{\name} & \textbf{85.8} & \textbf{86.6} & \textbf{84.0} & \textbf{80.1} & \textbf{83.8} & \textbf{75.6} & \textbf{79.3} & \textbf{80.3}  & \textbf{61.9} & \textbf{65.2}\\

    \bottomrule[1.1pt]
    \end{tabular}
}

\label{tab:ref}
\end{table*}

\noindent \textbf{Segmentation decoder.} The segmentation decoder $F_{decoder}$ predicts the masks $m$ and the corresponding mask scores $s$  based on three inputs: the fine-grained visual features $f_{img}$ from the frozen visual encoder, multi-granularity text embeddings $E_{v}$ from our VMTF module, and the mask embeddings $E_m$ from the output of LLM, following the similar decoding process~\cite{cheng2022masked, gu2024dataseg}. Formally,
\begin{equation}
    \{m_{j}, s_{j}, \}_{j=1}^{N} = F_{decoder}(f_{img}, E_{v}, E_m).
\end{equation}
where $m_{j}\in \mathbb{R}^{H\times W}$ is the j-th segmentation mask, $s_{j}\in \mathbb{R}$ denotes the mask scores of $m_{j}$, $N$ denotes $N$ mask proposals corresponding to the pre-defined $N$ mask tokens.
In practice, the final mask for each language instruction is derived by applying a threshold to the mask scores $s_{j}$.

\noindent \textbf{Training objectives.} 
We train \name~in an end-to-end manner on multiple tasks and datasets concurrently, using a unified loss function $\mathcal{L}$. 
Specifically, we employ the text loss $\mathcal{L}_{t}$ for text generation, the class loss $\mathcal{L}_{cls}$ for mask classification, and the mask loss $\mathcal{L}_{mask}$ for mask supervision. Formally,
\begin{equation}
\begin{aligned}
    \mathcal{L}= \mathcal{L}_{t}+&\lambda_{cls}\mathcal{L}_{cls}+\lambda_{mask}\mathcal{L}_{mask},\\
    \mathcal{L}_{mask}=&\lambda_{b}\mathcal{L}_{b}+\lambda_{d}\mathcal{L}_{d}.
\end{aligned}
\end{equation}
Specifically, we use an autoregressive cross-entropy loss for $\mathcal{L}_{t}$, a cross-entropy loss for $\mathcal{L}_{cls}$, and a combination of per-pixel binary cross-entropy loss $\mathcal{L}_{b}$ and DICE loss $\mathcal{L}_{d}$ for $\mathcal{L}_{mask}$. $\lambda$ denotes the weight of each loss component.

\begin{table*}[t]
  \centering
  \caption{ Comparison with the state-of-the-art models on more complex and challenging reasoning video object segmentation benchmark, ReVOS. Our \name~outperforms all the previous VLLM-based models with only 3B parameters.
  }
  \scalebox{0.98}{
    \begin{tabular}{l|c|ccc|ccc|ccc}
    \toprule[1.1pt] 
    \multirow{2}{*}{ Method } & \multirow{2}{*}{ Backbone } & \multicolumn{3}{c|}{ Reasoning } & \multicolumn{3}{c|}{ Referring } & \multicolumn{3}{c}{ Overall } \\
    \cline { 3 - 11 } & & $\mathcal{J}$ & $\mathcal{F}$ &  $\mathcal{J\&F}$ & $\mathcal{J}$ & $\mathcal{F}$ &  $\mathcal{J\&F}$ & $\mathcal{J}$ & $\mathcal{F}$ &  $\mathcal{J\&F}$ \\
    \midrule[0.7pt]
    
    LMPM~\cite{ding2023mevis}& Swin-T &  13.3 &24.3 &18.8 & 29.0 &39.1 &34.1 &  21.2 &31.7 &26.4 \\
    LLaMA-VID+LMPM~\cite{li2025llama}& Swin-T &  12.8 &23.7 & 18.2 & 29.0 &39.1 &34.1 &   20.9 &31.4 &26.1 \\
    ReferFormer~\cite{wu2022language}& Video-Swin-B  &  21.3 &25.6 & 23.4 & 31.2 &34.3 &32.7 &    26.2 &29.9 &28.1 \\
    \midrule[0.5pt] 
     LISA~\cite{Lai2023LISARS} & LLaVA-7B & 33.8 & 38.4 & 36.1 & 44.3 & 47.1 & 45.7 &  39.1 & 42.7 & 40.9 \\
    TrackGPT ~\cite{stroh2024trackgpt} & LLaVA-7B & 36.8 & 41.2 & 39.0  & 46.7  & 49.7  & 48.2   & 41.8  & 45.5  & 43.6 \\
    TrackGPT ~\cite{stroh2024trackgpt} & LLaVA-13B & 38.1  & 42.9 & 40.5  & 48.3  & 50.6  & 49.5   & 43.2  & 46.8  & 45.0 \\
    VISA ~\cite{yan2024visa} & Chat-UniVi-7B & 36.7  & 41.7 & 39.2  & 51.1  & 54.7  & 52.9   & 43.9  & 48.2  & 46.1 \\
    VISA ~\cite{yan2024visa} & Chat-UniVi-13B & 38.3  & 43.5 & 40.9  & 52.3  & 55.8  & 54.1   & 45.3  & 49.7  & 47.5 \\

    \midrule[0.5pt] 
   
    \textbf{\name} & Mipha-3B & \textbf{49.2} & \textbf{54.7} & \textbf{51.9} & \textbf{54.8} & \textbf{59.2} & \textbf{57.0} & \textbf{52.0} & \textbf{56.9}  & \textbf{54.5} \\

    \bottomrule[1.1pt]
    \end{tabular}
}

\label{tab:reason}
\end{table*}

\begin{table*}[t]
  \centering
  \caption{ Results of referring video object segmentation benchmarks, including Ref-YouTube-VOS val set and Ref-DAVIS17 val set.
  }
  \scalebox{0.97}{
    \begin{tabular}{c|l|c|ccc|ccc}
    \toprule[1.1pt] 
    \multirow{2}{*}{ Type } &
    \multirow{2}{*}{ Method } & 
    \multirow{2}{*}{ Backbone } & 
    \multicolumn{3}{c|}{  Ref-YouTube-VOS val } & \multicolumn{3}{c}{ Ref-DAVIS17 val }  \\
    
    \cline { 4 - 9 } & & & $\mathcal{J}$& $\mathcal{F}$ &$\mathcal{J\&F}$ &$\mathcal{J}$ & $\mathcal{F}$ & $\mathcal{J\&F}$  \\
    \midrule[0.7pt]
    \multirow{8}{*}{\shortstack{Segmentation\\Specialists}}
    & URVOS~\cite{seo2020urvos} & ResNet50 &  45.3 & 49.2 & 47.2 & 47.3 & 56.0 & 51.6 \\
    & YOFO~\cite{li2022you} & ResNet50 &  47.5 & 49.7 & 48.6 & 48.8 & 57.8 & 53.3 \\
    & MTTR\cite{botach2022end} & Video-Swin-T &  54.0 &56.6 &55.3 & - &- &- \\
    & ReferFormer~\cite{wu2022language}  & Video-Swin-B &  61.3 &64.6 &62.9& 58.1 &64.1 &61.1 \\
    & VLT~\cite{ding2022vlt} & Video-Swin-B & 63.8 &61.9 &65.6 & 61.6 &58.9 &64.3 \\
    & LMPM\cite{ding2023mevis} &Swin-T & -& -&- &- &- &-  \\
    & OnlineRefer~\cite{wu2023onlinerefer}  & Swin-L & 61.6 &65.5 &63.5 &61.6 &67.7 &64.8 \\
    & SgMg~\cite{miao2023spectrum}  & Video-Swin-B & 63.9 & 67.4 & 65.7 & 60.6 & 66.0 & 63.3 \\

    \midrule[0.5pt]
    \multirow{7}{*}{\shortstack{MLLM-based\\Segmentation Models}}
    & LISA~\cite{Lai2023LISARS} & LLaVA-7B & 53.4 &54.3 &53.9&  62.2 &67.3 & 64.8 \\
    & LISA~\cite{Lai2023LISARS} & LLaVA-13B & 54.0 &54.8 &54.4 & 63.2 &68.8 &66.0 \\
    & TrackGPT ~\cite{stroh2024trackgpt} & LLaVA-7B &  55.3 &57.4 &56.4 & 59.4 &67.0 &63.2 \\
    & TrackGPT ~\cite{stroh2024trackgpt} & LLaVA-13B & 58.1 &60.8 &59.5&62.7 &70.4 &66.5 \\
    & VISA ~\cite{yan2024visa} & Chat-UniVi-7B & 59.8 &63.2 & 61.5 & 66.3 &72.5 &69.4 \\
    & VISA ~\cite{yan2024visa} & Chat-UniVi-13B & 61.4 &64.7 &63.0 & 67.0 &73.8 &70.4 \\
    
    
     & \textbf{\name} & Mipha-3B & \textbf{65.4} & \textbf{69.5} & \textbf{67.5} & \textbf{67.3} & \textbf{74.9} & \textbf{71.1} \\
    
    \bottomrule[1.1pt]
    \end{tabular}
    }
  
\label{tab:exp-r-vos}
\end{table*}

\begin{table*}[t]
    \centering
    \caption{The performance comparison of our \name~on Multi-modal benchmarks. Our \name~presents promising performance compared with previous MLLMs in various question-answering benchmarks.}
    \scalebox{0.99}{
    \begin{tabular}{l|c|c|c|c|c|c}
    \toprule[1.1pt]  
    Method & 
    LLM & 
    VQA\textsuperscript{v2} & MMB  &  GQA & POPE & SQA  \\
    \midrule[0.5pt]

    BLIP-2~\cite{li2023blip} & Vicuna-13B & 65.0 & - & 41.0 & 85.3 & 61.0 \\
    InstructBLIP~\cite{instructblip} & Vicuna-7B &  - & 36.0 & 49.2 & - & 60.5  \\
    Shikra~\cite{chen2023shikra} & Vicuna-13B & 77.4 & 58.8 & - & - & -  \\
    Qwen-VL-Chat~\cite{bai2023qwen} & Qwen-7B  & 78.2 & 60.6 & 57.5 & - & 68.2  \\
    LLaVA-1.5~\cite{liu2024visual} & Vicuna-7B  & 78.5 & 64.3 & 62.0 & 85.9 & 66.8  \\
    MobileVLM~\cite{chu2023mobilevlm} & M-LLaMA-2.7B  & - & 59.6 & 59.0 & 84.9 & 61.0 \\ 

    \midrule[0.5pt]

    \name & Phi-2-2.7B  & 79.0 & 68.6 & 62.3 & 87.0 & 69.2 \\

    \bottomrule[1.1pt]
    \end{tabular}
    }
    
    \label{tab:mllm_results}
\end{table*}

\section{Experiments}
\textbf{Datasets.}
We train \name~in an end-to-end manner across multiple datasets of IVS tasks. For referring segmentation tasks, we use RefCOCO/+/g~\cite{yu2016modeling,nagaraja2016modeling} 
for image-level perception and Ref-Youtube-VOS~\cite{seo2020urvos} for video-level referring segmentation.
As for reasoning segmentation tasks, we use ReasonSeg~\cite{Lai2023LISARS} for image reasoning
segmentation and ReVOS~\cite{yan2024visa} for video-level reasoning and understanding.
Furthermore, we use LLAVA-150k~\cite{liu2024visual} for the vision-language instruction task following~\cite{Lai2023LISARS, zhang2024psalm}.

\noindent \textbf{Evaluation metrics.}
We report results in widely used evaluation metrics: cumulative Intersection-over-Union (cIoU) for referring expression segmentation (RES) task, cIoU and the average of all per-image Intersection-over-Unions (gIoU) for image reasoning segmentation task, and region similarity $\mathcal{J}$ and contour accuracy $\mathcal{F}$ for all the video-level segmentation tasks.

\noindent \textbf{Implementation details.}
We leverage the pre-trained light-weight Multi-modal Large Language Model Mipha-3B~\cite{zhu2024comprehensive} as our base model, Swin-B~\cite{liu2021swin} as our visual encoder and Maks2Former~\cite{cheng2022masked} as our segmentation decoder. All the models above are initialized with pre-trained weights. The layer numbers $N_1$ in OVP module and $N_2$ in VMTF are set to 3 by default.
We train \name~for 80k iterations with a batch size of 32 on 8×A100 GPUs. The total iterations are reduced to 40k in the ablation studies. By default, we set all the hyper-parameters $\lambda_{cls}$, $\lambda_{mask}$, $\lambda_{b}$, and $\lambda_{d}$ to 1.0 experimentally.

\subsection{Main Results}
We evaluate the effectiveness of the proposed \name~through the comprehensive comparison with other state-of-the-art methods on various IVS tasks and Multi-modal Question Answering tasks.

\noindent\textbf{Referring expression segmentation and reasoning segmentation.}
We compare \name~with the state-of-the-art methods on refCOCO/+/g~\cite{yu2016modeling,nagaraja2016modeling} and ReasonSeg~\cite{Lai2023LISARS} in \cref{tab:ref}. All the benchmarks are for image-level referring and reasoning segmentation.
Our end-to-end model with proposed modules, \name, surpasses all the previous works, achieving state-of-the-art performance on all the referring and reasoning datasets. To be specific, \name~surpasses the current SOTA by a large margin, reaching 80.1 cIoU on RefCOCO+ val (+7.2 over PSALM), 79.3 cIoU on RefCOCOg val (+5.5 over PSALM) and 61.9 gIoU on ReasonSeg (+9.0 over LISA-7B).

\noindent\textbf{Reasoning video object segmentation.}
We compare our \name~with the state-of-the-art methods on the challenging ReVOS~\cite{yan2024visa} benchmark in \cref{tab:reason}. Due to the dedicated and simplistic design, our 3B model beats the current SOTA with fewer parameters, especially in the ReVOS reasoning part, which demonstrates the effectiveness and efficiency of our \name. Specifically, we achieve 51.9 $\mathcal{J\&F}$ on reasoning, 57.0 $\mathcal{J\&F}$ on referring, and 54.5 $\mathcal{J\&F}$ on overall, which outperform previous SOTA model VISA-13B\cite{yan2024visa} 11.0, 2.9, and 7.0 respectively.

\noindent\textbf{Referring video object segmentation.}
We adopt two standard benchmarks for referring video object segmentation tasks, including Ref-YouTube-VOS and Ref-DAVIS17.
The comparisons between \name~with the previous state-of-the-art R-VOS methods are demonstrated in \cref{tab:exp-r-vos}.
\name~surpasses all the segmentation specialists and MLLM-based segmentation models on both benchmarks. 
In challenging Ref-YouTube-VOS, our method achieves the best performance 67.5 on $\mathcal{J\&F}$ (+4.5 over VISA), outperforming all the methods by a large margin. 
Besides, \name~achieves remarkable performance with only 3B parameters among MLLM-based models like TrackGPT-13B and VISA-13B.

\noindent\textbf{Multi-modal question answering benchmarks.}
Our \name~is the unified segmentation model leveraging Multi-modal Large Language Models (MLLMs) for referring and reasoning segmentation in image and video domains, which has powerful reasoning capabilities for detailed text instructions.
We also demonstrate its efficacy in addressing vision-language understanding tasks.
As illustrated in \cref{tab:mllm_results}, we evaluate \name~on various Multi-modal benchmarks. 
Notably, \name~achieves commendable performance with a smaller model size compared to existing MLLMs, such as BLIP-2~\cite{li2023blip}, Qwen-VL-Chat~\cite{bai2023qwen}, and LLaVA-1.5~\cite{liu2024visual}.
These results underscore the model's advanced conversational and reasoning capabilities.

\subsection{Ablations}

\noindent\textbf{Effectiveness of the proposed components.} We assess the effectiveness of the proposed OVP module and VMTF module.
As shown in \cref{tab:ab-component}, with our object-aware video perceiver and vision-guided multi-granularity text fusion, the segmentation accuracy can be promoted significantly on both referring and reasoning tasks, which demonstrates the effectiveness of our distinct design for each module.

\begin{table}[t]
  \centering
  \caption{ Ablation on the core components of \name. 
  OVP and VMTF denote the proposed Object-aware Video Perceiver and Vision-guided Multi-granularity Text Fusion module.}
  \scalebox{0.75}{
    \begin{tabular}{c|c|c|ccc}
    \toprule[1.1pt] 
    \multirow{2}{*}{ OVP} & 
    \multirow{2}{*}{ VMTF } & 
    \multicolumn{1}{c|}{  RefCOCO val } &
    \multicolumn{3}{c}{  ReVOS }  \\
    \cline { 3 -6 }  & &cIoU & Reasoning & Referring & Overall  \\
    \midrule[0.7pt]
    
    &  & 83.3 & 50.2 & 55.7 & 52.9 \\
     \Checkmark &  & 84.6 & 51.2 & 56.9 & 54.0 \\
     & \Checkmark & 84.8 & 51.3  & 56.6 & 53.9 \\ 
     \Checkmark & \Checkmark & \textbf{85.8} & \textbf{51.9} & \textbf{57.0} & \textbf{54.5} \\

    \bottomrule[1.1pt]
    \end{tabular}
    }

    \vspace{-2mm}
  
\label{tab:ab-component}
\end{table}

\noindent\textbf{Different reference frames number.}
We evaluate the performance of different numbers of reference frames $T_r$ in our object-aware video perceiver module.
As shown in \cref{tab:ab-number}, the performance of our \name~gradually improves as the number $T_r$ increases. We adopt $T_r$=4 for effective and efficient training and inference.

\begin{table}[t]
  \centering
  \caption{ The performance comparison on video segmentation tasks with different number $T_r$ of reference frames in our object-aware video perceiver module.
  }
  \scalebox{0.8}{
    \begin{tabular}{c|ccc|ccc}
    \toprule[1.1pt] 
    \multirow{2}{*}{ $T_r$ } & 
    \multicolumn{3}{c|}{  Ref-DAVIS17 } &
    \multicolumn{3}{c}{  ReVOS }  \\
    \cline { 2 - 7 } & $\mathcal{J}$& $\mathcal{F}$ &$\mathcal{J\&F}$ &  Reasoning & Referring & Overall \\
    \midrule[0.7pt]
    0 & 66.4 & 74.0 & 70.2 & 50.9 & 56.4 & 53.6  \\
   4 & 67.3 & 74.9 & 71.1 & 51.9 & 57.0 & 54.5  \\
   8 & 67.6 & 75.0 & 71.3 & 52.0 & 57.2 & 54.6  \\

    \bottomrule[1.1pt]
    \end{tabular}
    }
  \vspace{-1mm}
  
\label{tab:ab-number}
\end{table}

\noindent\textbf{Different text fusion strategy.}
In \cref{tab:ab-vmtf}, we compare our vision-guided multi-granularity text fusion module with various design choices.
The comparison of the results shows that our multi-granularity fusion strategy (global + detailed) outperforms the single mode.
This highlights the distinct properties of Instructed Visual Segmentation tasks, where both global and detailed textual information are crucial for comprehensive reasoning and perception.

\begin{table}[t]
  \centering
  \caption{ Ablation of different text fusion strategy.
  }
  \scalebox{0.73}{
    \begin{tabular}{c|c|c|ccc}
    \toprule[1.1pt] 
    \multirow{2}{*}{ Global} & 
    \multirow{2}{*}{ Detailed } & 
    \multicolumn{1}{c|}{  RefCOCO val } &
    \multicolumn{3}{c}{  ReVOS }  \\
    \cline { 3 -6 }  & &cIoU & Reasoning & Referring & Overall  \\
    \midrule[0.7pt]
    
     \Checkmark &  & 84.2 & 50.3 & 56.7 & 53.5 \\
     & \Checkmark & 83.5 & 51.6 & 56.7 & 54.1 \\
     \Checkmark & \Checkmark & \textbf{85.8} & \textbf{51.9} & \textbf{57.0} & \textbf{54.5} \\

    \bottomrule[1.1pt]
    \end{tabular}
    }
  \vspace{-3mm}
\label{tab:ab-vmtf}
\end{table}

\noindent\textbf{Comparison between different training data recipes.} 
Our model achieves excellent performance on all IVS tasks with joint training.
In ~\cref{tab:ab-data}, we evaluate the impact of different training data recipes on our model performance. 
For different visual types, the model trained on images exhibits excellent zero-shot performance on ReasonVOS, demonstrating the generalization and robustness of our \name. Similarly, the model trained on videos achieves promising performance on RefCOCO, even surpassing non-zero-shot methods like LISA and VISA finetuning on the RefCOCO datasets.
For different task types, the model trained exclusively on reasoning datasets delivers promising performance on RefCOCO, surpassing specialist models like ReLA and PolyFormer. This demonstrates that our \name~with powerful reasoning segmentation capabilities can effectively generalize to referring segmentation tasks.

\begin{table*}[t]
  \centering
  \caption{
  The performance comparison between different visual types (image and video) and task types (reasoning and referring) of Instructed Visual Segmentation tasks. Training jointly enables \name~to achieve better performance with a single model.
  }
  \scalebox{0.87}{
    \begin{tabular}{c|c|c|c|ccc|ccc|ccc}
    \toprule[1.1pt] 
    \multirow{2}{*}{Image} & 
    \multirow{2}{*}{ Video } &
    \multirow{2}{*}{Reasoning} & 
    \multirow{2}{*}{ Referring } &
    \multicolumn{3}{c|}{  RefCOCO } &
    \multicolumn{3}{c|}{  RefCOCO+ } & \multicolumn{3}{c}{ ReVOS }   \\
    \cline { 5 - 13 } & & & & val & testA & testB & val & testA & testB & Reasoning & Referring & Overall  \\
    \midrule[0.7pt]
    \Checkmark & & \Checkmark & \Checkmark &  84.2 & 86.0 & 81.7 & 79.2 & 83.3 & 72.9 & 43.6 & 53.9 & 48.7  \\
    & \Checkmark & \Checkmark & \Checkmark &  78.1 & 82.1 & 75.0 & 70.0 & 77.6 & 63.0 & 50.0 & 56.7 & 53.3  \\
    \Checkmark & \Checkmark & \Checkmark &   & 74.0 & 78.0 & 69.6 & 65.3 & 72.6 & 57.9 & 51.6 & 55.9 & 53.8  \\
    \Checkmark & \Checkmark & & \Checkmark &  85.1 & 86.6 & 83.2 & 80.9 & 84.2 & 76.1 & 37.4 & 52.2 & 44.8  \\
    
     \Checkmark & \Checkmark & \Checkmark & \Checkmark & \textbf{85.8} & \textbf{86.6} & \textbf{84.0} & \textbf{80.1} & \textbf{83.8} & \textbf{75.6} & \textbf{51.9} & \textbf{57.0} & \textbf{54.5} \\ 
    
    \bottomrule[1.1pt]
    \end{tabular}
    }

    \vspace{-1mm}
  
\label{tab:ab-data}
\end{table*}

\section{Conclusion}
In this work, we unify the text-guided segmentation tasks across image and video domains under the framework of Instructed Visual Segmentation (IVS) and introduce \name, a universal and simplified segmentation network for IVS tasks. Our approach reduces the complexity and overcomes challenges associated with unifying referring expressions segmentation and reasoning segmentation at both the image and video levels. We present two main designs: an object-aware video perceiver that effectively captures temporal and object-specific information for enhanced video understanding and a vision-guided multi-granularity text fusion module that seamlessly integrates global and detailed language instructions with fine-grained vision guidance. Through multi-task and end-to-end training, \name~achieves superior performance in all IVS tasks, surpassing both specialist models and existing MLLM-based methods. 

{
    \small
    \bibliographystyle{ieeenat_fullname}
    \bibliography{main}
}

\clearpage
\setcounter{page}{1}
\maketitlesupplementary
\appendix

\section{Additional Implementation Details}
\label{sec:implement_details}

We utilize Phi-2~\cite{javaheripi2023phi} with 2.7B parameters as our Large Language Model,  SigLIP~\cite{zhai2023sigmoid} as our CLIP Encoder, Swin-B~\cite{liu2021swin} as our Visual Encoder, and pre-trained Maks2Former~\cite{cheng2022masked} as our Segmentation Decoder.
We keep both CLIP Encoder and Visual Encoder frozen while applying LORA with a rank of 8 to fine-tune the Large Language Model.
In contrast, the OVP, VMTF, and Segmentation Decoder components are fully fine-tuned. 
We use AdamW optimizer with the learning rate and weight decay set to 0.00004 and 0, respectively. In addition, we adopt Cosine Decay for the learning rate schedule, where the warmup steps are set to 1680.
The source code for our implementation will be made publicly available in the near future.

\section{Segmentation Decoder Structure}

\begin{figure}[t]
    \centering
    \includegraphics[width=0.48\textwidth]{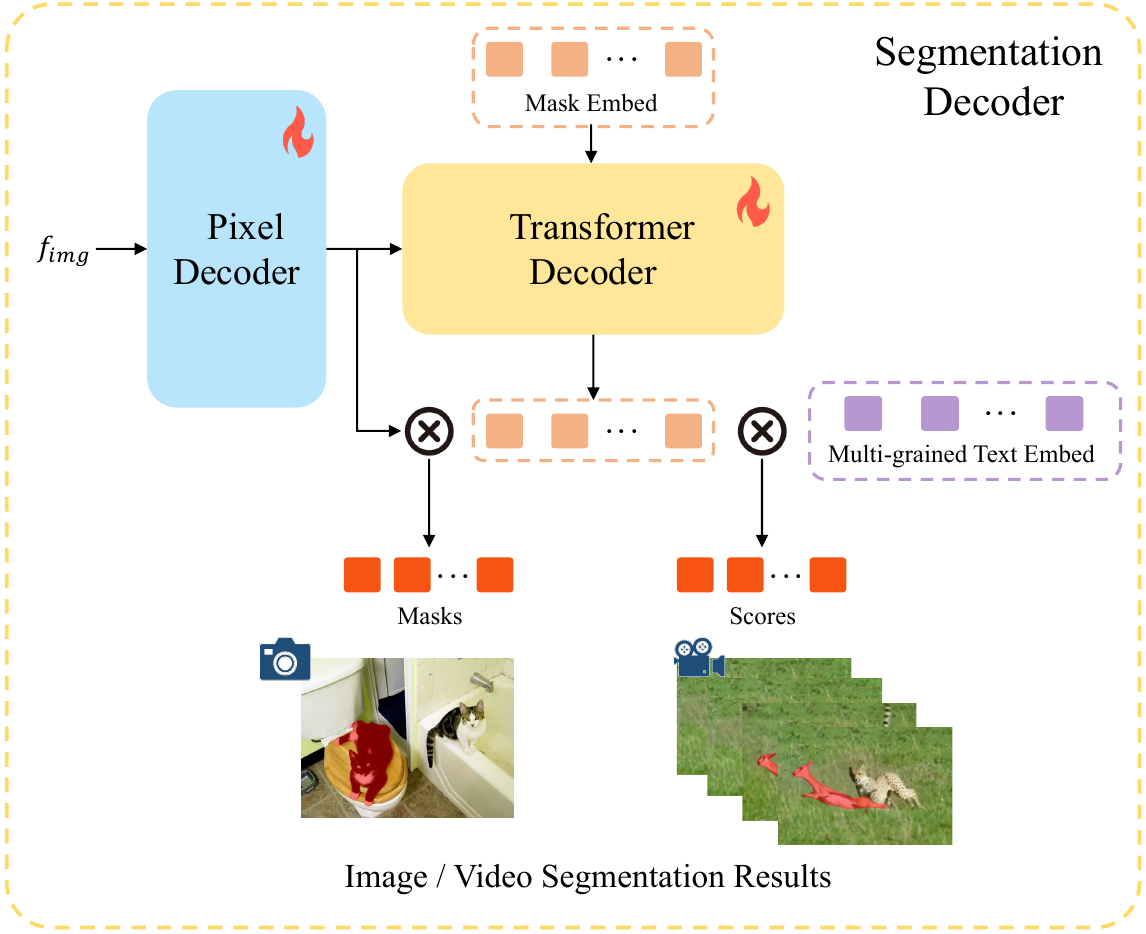}
    \caption{The structure of the Segmentation Decoder module. Following ~\cite{cheng2022masked}, we adopt the pixel decoder and transformer decoder to excavate pixel-level visual information and instance-level object information. In contrast, we calculate the similarity between mask embeddings and multi-grained text embeddings as the mask scores for mask proposals' selection.
    }
    \label{fig:decoder}
\end{figure}

We illustrate the architecture of the segmentation decoder module in Fig. \ref{fig:decoder}. Consistent with previous methods~\cite{cheng2022masked,zhang2024psalm}, our approach integrates both a pixel decoder and a transformer decoder to extract pixel-level visual information and instance-level object information. Distinctively, we compute the similarity between mask embeddings and multi-grained text embeddings to derive mask scores, which are then utilized for the selection of mask proposals.

\begin{table*}[h]
    \centering
    \caption{Task-specific language instructions for all the Instructed Visual Segmentation tasks..}
    \scalebox{0.63}{
    \begin{tabular}{l|c|c|c|c}
    \toprule[1.1pt]  
    Task & Visual Type &
    Dataset & Instruction Template & Text Prompt \\
    \midrule[0.5pt]

\multirow{2}{*}{Referring Expression Segmentation}              & \multirow{2}{*}{Image}  & \multirow{2}{*}{RefCOCO/+/g}                             & \multirow{2}{*}{\textit{\makecell[c]{You need to perform Referring Expression Segmentation \\ on the image according to the Text Prompt.}}} & \multirow{2}{*}{"A baseball catcher with an open mitt"}              \\
    &   &   &         \\
& & & \\

\multirow{2}{*}{Reasoning Segmentation}              & \multirow{2}{*}{Image}  & \multirow{2}{*}{ReasonSeg}                             & \multirow{2}{*}{\textit{\makecell[c]{You need to perform Reasoning Segmentation \\ on the image according to the Text Prompt.}}} & \multirow{2}{*}{\makecell[c]{"The person who appears to have \\ already won in the battle"}}       \\
    &   &   &         \\
& & & \\

\multirow{2}{*}{Referring Video Object Segmentation }              & \multirow{2}{*}{Video}  & \multirow{2}{*}{Ref-YouTube-VOS, etc.}                             & \multirow{2}{*}{\textit{\makecell[c]{You need to perform Referring Video Object Segmentation  \\ on the video according to the Text Prompt.}}} & \multirow{2}{*}{"A duck is held by a person with her both hands"}             \\
    &   &   &         \\
& & & \\

\multirow{2}{*}{Reasoning Video Object Segmentation}              & \multirow{2}{*}{Video}  & \multirow{2}{*}{ReVOS}                             & \multirow{2}{*}{\textit{\makecell[c]{You need to perform Reasoning Video Object Segmentation \\ on the video according to the Text Prompt.}}} & \multirow{2}{*}{"Which person is in the leading position?"}             \\
    &   &   &         \\
& & & \\

    \bottomrule[1.1pt]
    \end{tabular}
    }
    
    \label{tab:prompt_design}
\end{table*}

\begin{figure*}[t]
    \centering
    \includegraphics[width=\textwidth]{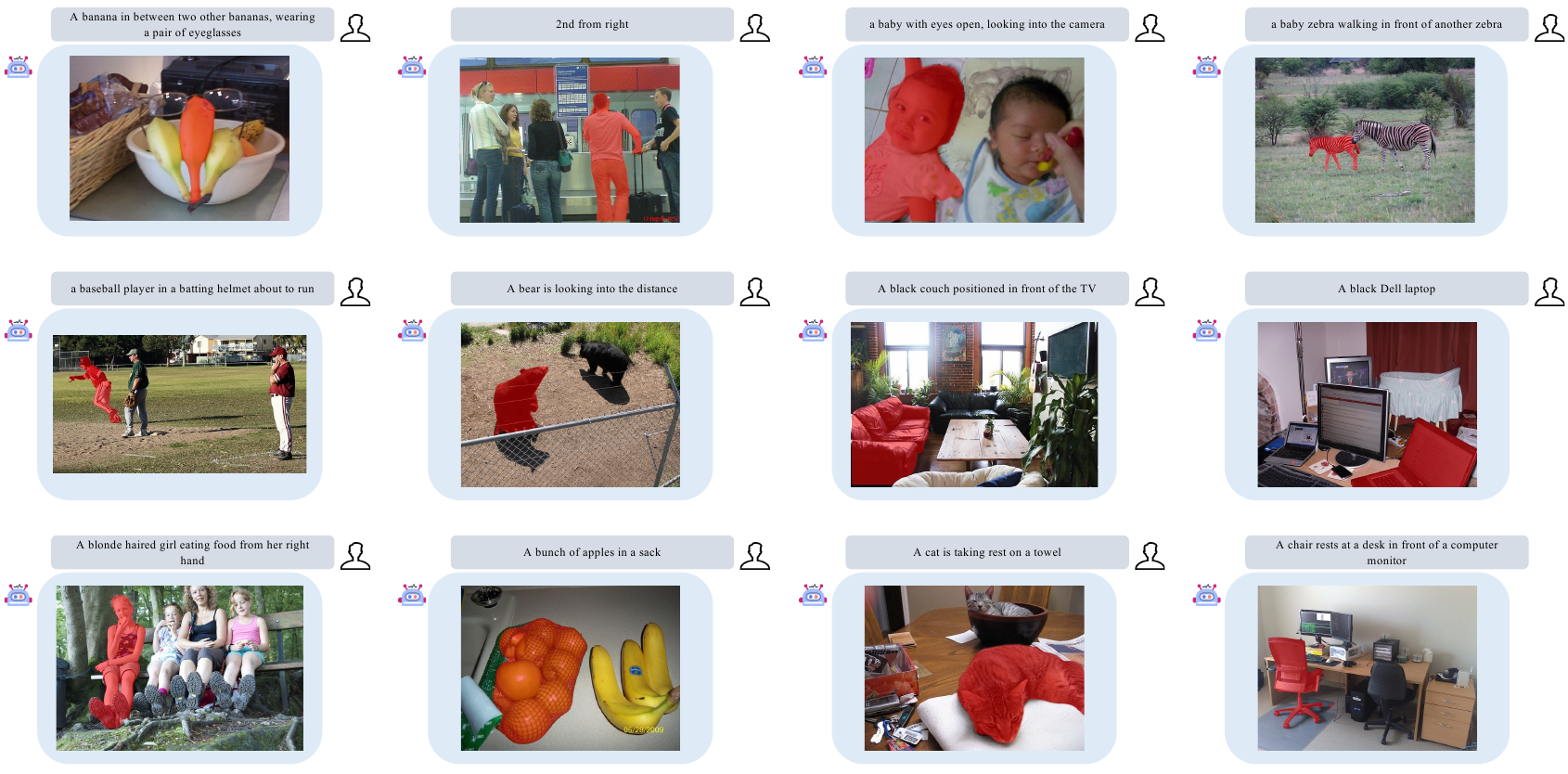}
    \caption{Qualitative results of \name’s capability in referring expression segmentation. 
    }
    \label{fig:visualize_res}
\end{figure*}

\section{Task-specific Instructions Design}

In this section, we illustrate the text prompt with task-specific instructions for all the Instructed Visual Segmentation tasks. As shown in Tab.~\ref{tab:prompt_design}, we design different instruction templates for all four segmentation tasks along with corresponding text prompts.

\section{More Qualitative Results}

\subsection{Referring Expression Segmentation (Image-level)}
Fig. \ref{fig:visualize_res} shows the visualization of \name~on referring segmentation task.

\subsection{Reasoning Segmentation (Image-level)}
\begin{figure*}[t]
    \centering
    \includegraphics[width=\textwidth]{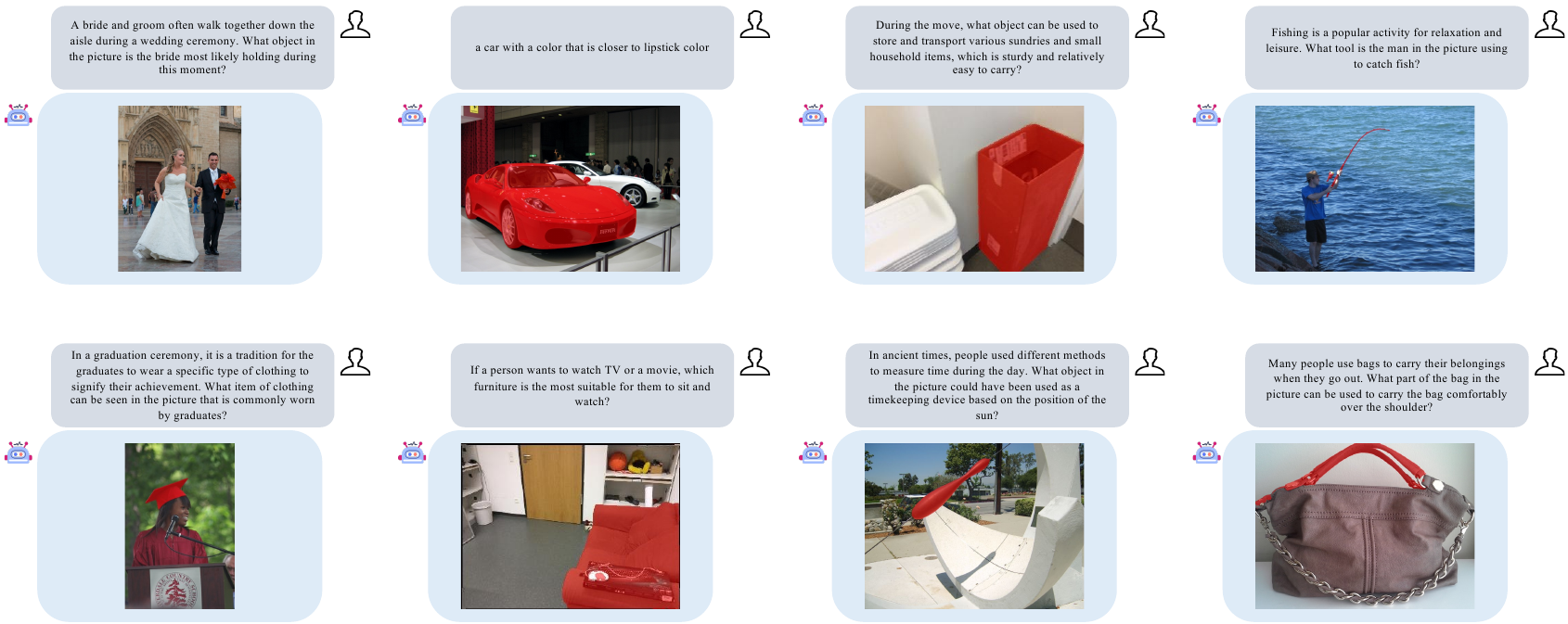}
    \caption{Qualitative results of \name~in 
    reasoning segmentation.
    }
    \label{fig:visualize_reason}
\end{figure*}

Fig. \ref{fig:visualize_reason} presents the effectiveness of our \name~in understanding the complex question and performing segmentation according to the reasoning process.

\subsection{Reasoning and Referring Video Object Segmentation (Video-level)}

\begin{figure*}[t]
    \centering
    \includegraphics[width=\textwidth]{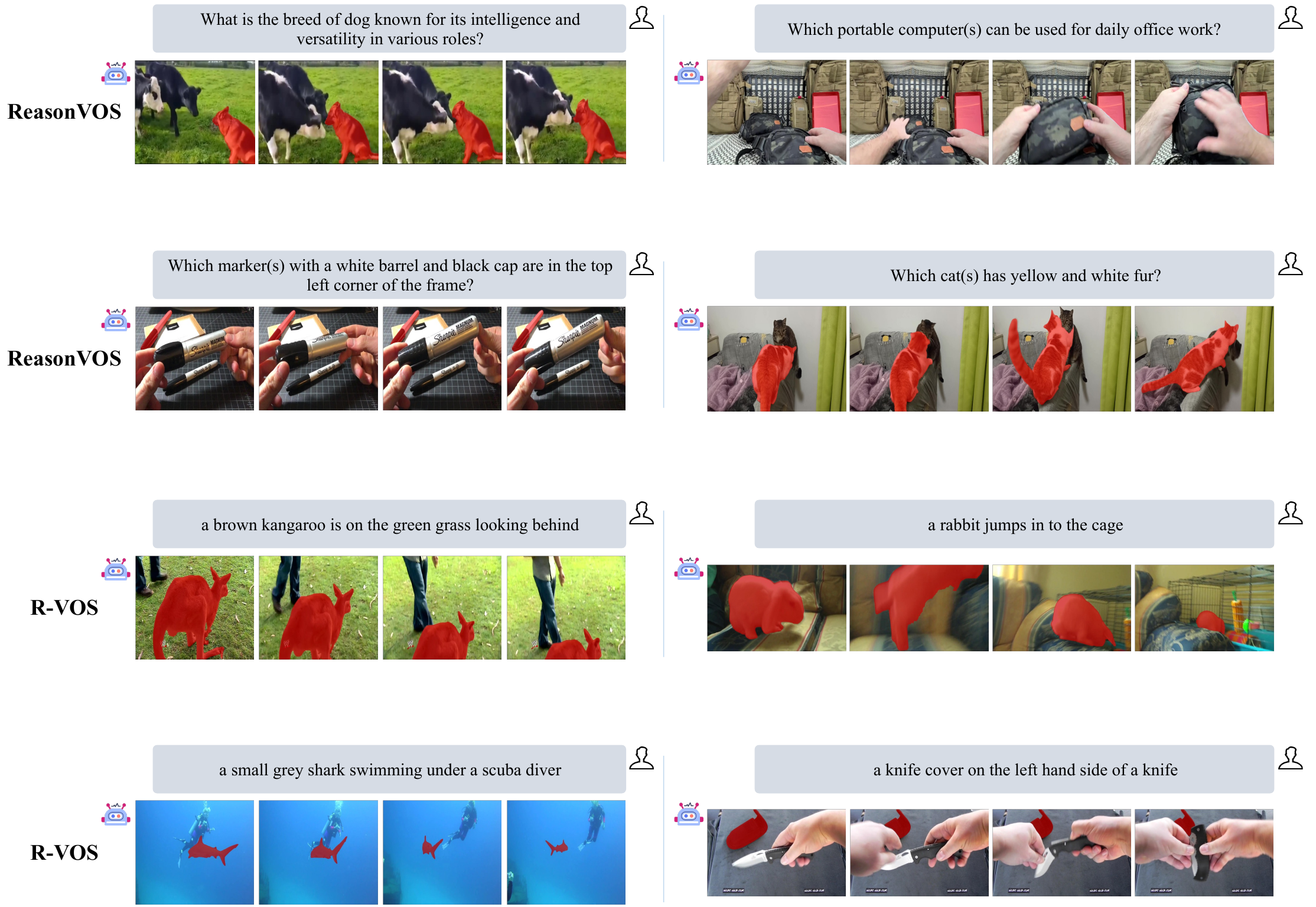}
    \caption{Qualitative results of \name~demonstrate its capability in the complex reasoning video object segmentation task and referring video object segmentation task.
    }
    \label{fig:visualize_revos}
\end{figure*}

Fig. \ref{fig:visualize_revos} shows the effectiveness of \name~in comprehending both the reasoning questions and temporal coherence. \name~is capable of producing segmentation masks that maintain consistency across temporal sequences.

\end{document}